%% file: icmr_paper.tex
\documentclass[sigconf,authorversion]{acmart}

\AtBeginDocument{%
  \providecommand\BibTeX{{%
    \normalfont B\kern-0.5em{\scshape i\kern-0.25em b}\kern-0.8em\TeX}}}

\copyrightyear{2022}
\acmYear{2022}
\setcopyright{acmlicensed}\acmConference[ICMR '22]{Proceedings of the 2022 International Conference on Multimedia Retrieval}{June 27--30, 2022}{Newark, NJ, USA}
\acmBooktitle{Proceedings of the 2022 International Conference on Multimedia Retrieval (ICMR '22), June 27--30, 2022, Newark, NJ, USA}
\acmPrice{15.00}
\acmDOI{10.1145/3512527.3531395}
\acmISBN{978-1-4503-9238-9/22/06}

\newcommand{\ie}{i.e. }
\newcommand{\eg}{e.g. }
\newcommand\R{\mathbb{R}}
\newcommand\rel{\mathcal{R}}
\usepackage{graphicx}
\usepackage{amsmath}
\usepackage{booktabs}
\usepackage[normalem]{ulem}
\usepackage{multirow}
\newcommand\rev[2]{{\sout{#1}}\textcolor{black}{#2}}
\newcommand{\new}[1]{{\color{black}#1}}
\definecolor{darkgreen}{RGB}{0,128,0}
\begin{document}
\fancyhead{}

\title{Relevance-based Margin for Contrastively-trained Video Retrieval Models}

\author{Alex Falcon}
\email{falcon.alex@spes.uniud.it}
\orcid{0000-0002-6325-9066}
\affiliation{%
  \institution{Fondazione Bruno Kessler}
  \streetaddress{Via Sommarive, 18}
  \city{Trento}
  %\state{Trento}
  \country{Italy}
  \postcode{38123}
}
\affiliation{%
  \institution{University of Udine}
  \streetaddress{Via delle Scienze, 206}
  \city{Udine}
  %\state{Udine}
  \country{Italy}
  \postcode{33100}
}

\author{Swathikiran Sudhakaran}
\email{swathikirans@gmail.com}
\orcid{0000-0002-7261-3581}
\affiliation{%
  \institution{Samsung AI Center Cambridge}
  \streetaddress{7th Floor, 50 Station Rd}
  \city{Cambridge}
  %\state{Trento}
  \country{United Kingdom}
  \postcode{CB12JH}
}

\author{Giuseppe Serra}
\email{giuseppe.serra@uniud.it}
\orcid{0000-0002-4269-4501}
\affiliation{%
  \institution{University of Udine}
  \streetaddress{Via delle Scienze, 206}
  \city{Udine}
  %\state{Udine}
  \country{Italy}
  \postcode{33100}
}

\author{Sergio Escalera}
\email{sergio@maia.ub.es}
\orcid{0000-0003-0617-8873}
\affiliation{%
  \institution{University of Barcelona and Computer Vision Center}
  \streetaddress{Gran Via de les Corts Catalanes 585}
  \city{Barcelona}
  %\state{Trento}
  \country{Spain}
  \postcode{08007}
}

\author{Oswald Lanz}
\email{lanz@inf.unibz.it}
\orcid{0000-0003-4793-4276}
\affiliation{%
  \institution{Free University of Bozen-Bolzano}
  \streetaddress{Piazza Domenicani, 3}
  \city{Bolzano}
  %\state{Bolzano}
  \country{Italy}
  \postcode{39100}
}
%\email{webmaster@marysville-ohio.com}
%\affiliation{%
%  \institution{Institute for Clarity in Documentation}
%  \streetaddress{P.O. Box 1212}
%  \country{USA}

\renewcommand{\shortauthors}{Author, et al.}

\begin{abstract}
   \input{sub/0_abstract}
\end{abstract}





\maketitle

\input{sub/1_introduction}

\input{sub/2_related}

\input{sub/3_method}

\input{sub/4_results}

\input{sub/5_discussion}

\begin{acks}
We gratefully acknowledge the support from Amazon AWS Machine Learning Research Awards (MLRA) and NVIDIA AI Technology Centre (NVAITC), EMEA. We acknowledge the CINECA award under the ISCRA initiative, which provided computing resources for this work. This work has been partially supported by the Spanish project PID2019-105093GB-I00 and by ICREA under the ICREA Academia programme.
\end{acks}

%\newpage
%%%%%%%%% REFERENCES
%{\small
\bibliographystyle{ACM-Reference-Format}
\bibliography{biblio}
%}

\input{sub/A_supplementary}

\end{document}

%% file: sub/0_abstract.tex
Video retrieval using natural language queries has attracted increasing interest due to its relevance in real-world applications, from intelligent access in private media galleries to web-scale video search. Learning the cross-similarity of video and text in a joint embedding space is the dominant approach. To do so, a contrastive loss is usually employed because it organizes the embedding space by putting similar items close and dissimilar items far.
This framework leads to competitive recall rates, as they solely focus on the rank of the groundtruth items. Yet, assessing the quality of the ranking list is of utmost importance when considering intelligent retrieval systems, since multiple items may share similar semantics, hence a high relevance. 
Moreover, the aforementioned framework uses a fixed margin to separate similar and dissimilar items, treating all non-groundtruth items as equally irrelevant. In this paper we propose to use a variable margin: we argue that varying the margin used during training based on how much relevant an item is to a given query, i.e.~a relevance-based margin, easily improves the quality of the ranking lists measured through nDCG and mAP. We demonstrate the advantages of our technique using different models on EPIC-Kitchens-100 and YouCook2. We show that even if we carefully tuned the fixed margin, our technique (which does not have the margin as a hyper-parameter) would still achieve better performance. Finally, extensive ablation studies and qualitative analysis support the robustness of our approach. Code will be released at \url{https://github.com/aranciokov/RelevanceMargin-ICMR22}. 

%% file: sub/1_introduction.tex
\section{Introduction}
\begin{figure}
    \centering
    \includegraphics[width=\linewidth]{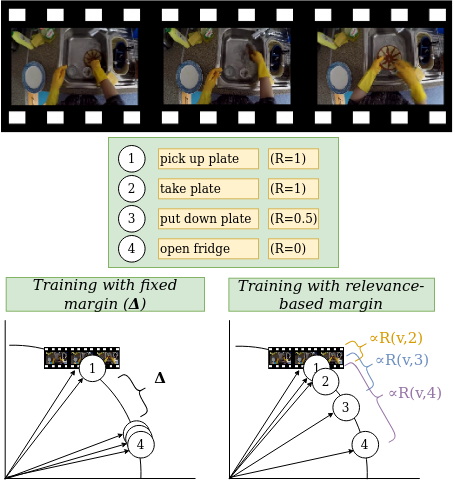}
    \caption{Training a model for text-video retrieval by employing a contrastive loss which uses a fixed margin $\Delta$ (lower left) treats semantically equivalent descriptions which do not appear as groundtruth pairs in the dataset as equally irrelevant. We propose to move away from such a paradigm and adopt a relevance-based margin (lower right), \ie a margin which is proportional to the relevance $\rel$ (see Sec.~\ref{sec:relevance}). } %\sergio{this figure should be cited within intro.}}
    \label{fig:method}
\end{figure}
With the rapid growth of digital media shared on the web it becomes increasingly important for real-world applications to offer flexible, user friendly modalities to access media content at scale. Google video search for example, translates a natural language query into a ranked list of content-related videos from the web. Natural free form, unrestricted language enables a user to express the fine-grained details in an articulated query, and each user can do so with its own expressivity. Thus, a same retrieval response can be triggered with syntactically different but semantically coherent queries. This poses significant challenges to the current state of the art in cross-modal retrieval research. 

Recent approaches which deal with cross-modal video retrieval aim at learning a joint embedding space \cite{Chen_2020_CVPR,croitoru2021teachtext,dong2021dual,wang2021t2vlad} by means of contrastive losses \cite{hadsell2006dimensionality,schroff2015facenet,miech2020end,oord2018representation}, which put the associations available in the dataset (\eg a video and its natural language description) as close as possible while enforcing a separation margin to all the other items (see lower left of Fig.~\ref{fig:method}). During inference, the ranking list for a given query is produced by computing similarity scores with respect to all the items by means of, \eg the dot product or the cosine similarity. %\swathi{not dot product always; depends on the similarity measure. generalize the statement} is used to rapidly compute a similarity score, which can then be used to produce the ranking list. 
By measuring the performance of the video retrieval system with rank-unaware metrics, such as recall rates, increasingly better solutions to this problem were proposed. In fact, contrastive losses synergize well with recall rates, given how they maximize the similarity of the associated items. But during training they do not make any distinction between items which are \textit{highly relevant} and items which are only \textit{partially} or \textit{completely irrelevant} to a given query. %While effective and powerful as a framework \sergio{While effective and powerful as a framework - rewrite in a more simple manner and also mention recall explicitely again}, during training it does not make any distinction between items which are \textit{highly relevant} and items which are only \textit{partially} or \textit{completely irrelevant} to a given query. 
For example, if a query is about `how to cook a pizza', then videos which depict how to `bake a pizza', `cook pasta', or `knead dough' are all treated the same way, although they can be more or less semantically close to the query. \rev{}{Furthermore, one of the reasons which limited the usage of rank-aware metrics in video retrieval consists in visual-language datasets only providing the visual contents and textual annotations (obtained manually \cite{xu2016msrvtt,zhou2018towards} or automatically \cite{miech2019howto100m}). Due to the absence of relevance grades, rank-aware metrics (\eg nDCG) are difficult to adopt. Recently, this problem was partially alleviated by the introduction of a relevance function \cite{damen2020rescaling} which, to avoid a costly manual annotation step, is defined in terms of the captions already available in the dataset.}

\rev{}{To give the model awareness of the semantical differences between items and queries during training, we free the margin from its stillness. Several solutions for non-fixed margins were proposed in previous literature, such as using multiple margins (e.g. \cite{cheng2016person}) or adaptive solutions. In particular, \cite{semedo2019cross} implemented a schedule for the margin value which gradually incorporates inter-category correlations and information about the structure of the embedding space. Recently, for video retrieval \cite{he2021improving} proposed an adaptive margin proportional to the similarity of item and query as computed by multiple models. Differently from them, we propose to inject semantic knowledge into the training process by means of a relevance-based margin.}
%One of the reasons which limited the usage of rank-aware metrics in video retrieval is due to the fact that visual-language datasets only provide the visual contents and textual annotations (obtained manually \cite{xu2016msrvtt,damen2020rescaling,zhou2018towards} or automatically \cite{miech2019howto100m}). Since they do not provide relevance grades, rank-aware metrics (\eg nDCG) are difficult to adopt. Recently, this problem was partially alleviated by the introduction of a relevance function \cite{damen2020rescaling} which, to avoid a costly manual annotation step, is defined in terms of the captions already available in the dataset. In this work, in place of the fixed margin which is commonly used in video retrieval.
To do so, we leverage the relevance function detailed in \cite{damen2020rescaling}, so that the margin is proportional to how relevant the item is to the query, as illustrated in Fig.~\ref{fig:method}. By doing so, we effectively discard one hyper-parameter to tune. Moreover, even by performing an expensive search for it, the results are still suboptimal when compared to the proposed relevance-based margin. We give empirical evidence that the proposed technique makes it possible to easily improve the quality of the ranking lists, measured through Normalized Discounted Cumulative Gain (nDCG) and Mean Average Precision (mAP). We use three different and increasingly more complex models (MME from \cite{wray2019fine}, JPoSE \cite{wray2019fine}, and HGR \cite{Chen_2020_CVPR}) on two datasets (EPIC-Kitchens-100 \cite{damen2020rescaling} and YouCook2 \cite{zhou2018towards}). Furthermore, we perform several ablations to study how it interacts with multiple video modalities (motion, appearance, audio) and with both cross-modal and within-modal losses. 

We organize the paper as follows. In Section \ref{sec:rw} we review related works, including vision and language tasks, main techniques and losses used to deal with text-video retrieval, and optimization of retrieval metrics such as the nDCG. Then, we formally describe the proposed technique in Sec.~\ref{sec:fm}, in terms of the relevance function and how we apply it to a typical contrastive loss setting. In Sec.~\ref{sec:exp} we perform multiple experiments to prove the strength of the relevance-based margin. Finally, in Sec.~\ref{sec:cs} we conclude the paper. %proposing some exciting future work.

%% file: sub/2_related.tex
\section{Related works\label{sec:rw}}
\textbf{Vision and Language.} %Being able to correctly match a natural language description and corresponding visual contents has been a research track for several decades (\eg text-based image retrieval \cite{chang1992image,tamura1984image}). In recent years, deep learning brought several advancements in multiple tasks dealing with images and language, such as question answering \cite{anderson2018bottom,antol2015vqa,yang2016stacked}, retrieval \cite{lee2021cosmo,zhang2020context}, and captioning \cite{shi2021enhancing,dong2021captioning}. Increasing attention has also been directed towards the video counterparts, leading to more challenging environments since both spatial and temporal reasoning are needed, \eg for video question answering \cite{huang2020location,kim2020modality,li2020hero,yang2020bert}, captioning \cite{lei2020mart,li2020hero}, and text-to-video retrieval \cite{Chen_2020_CVPR,dong2021dual,gabeur2020multi,liu2019use,miech2019howto100m,mithun2018learning,yu2018joint}. Moreover, %several opportunities have also been raised thanks to the multimodal nature of videos, \eg audio tracks and automatically transcripted narrations \cite{alayrac2020self,miech2019howto100m,mithun2018learning,nagrani2020speech2action,rouditchenko2020avlnet,sun2019videobert}, raised several opportunities. In this work, we focus on two symmetric versions of video retrieval, that is text-to-video and video-to-text retrieval.
%\textcolor{red}{swathi: suggestion. in recent years, deep learning brought... dealing with vision and language, such as question answering(image citation1, image citation2, video citation1, video citaton2), retrieval(imag1, imag2, video1, video2), captioning(image1, image2, video1, video2). mention about methods leveraging textual data for vision model pretraining(virtex, clip, align). mention methods that pretrain both vision and textual models jointly (lxmert, visualbert, videobert, uniter, univlm, oscar, etc: pick 2 each for image and video). these perform very well for downstream tasks (captioning, vqa, retrieval) but are data hungry and expensive to train, impractical from a computational point of view}
In recent years, deep learning brought several advancements in multiple tasks dealing with vision and language, such as question answering \cite{anderson2018bottom,antol2015vqa,huang2020location,kim2020modality}, retrieval \cite{lee2021cosmo,zhang2020context,dong2021dual,Chen_2020_CVPR}, and captioning \cite{shi2021enhancing,dong2021captioning,lei2020mart,li2020hero}. Given that vast amounts of data can be scraped from the web, many works perform a joint vision and language pretraining \cite{li2020oscar,chen2020uniter,sun2019videobert,zhou2021cupid} by optimizing vision-text proxy tasks. Recently, a line of research uses natural language supervision such as captioning~\cite{desai2021virtex} or alignment~\cite{jia2021scaling} objectives to pretrain visual models. While in both cases they achieve competitive and state-of-the-art results on downstream tasks, these methods are data hungry and expensive to train, making them impractical from a computational point of view. 

\textbf{Text-Video Retrieval.} Multiple techniques were proposed to learn a representation for the input data while capturing multimodal interactions. \cite{liu2019use,wang2021t2vlad,gabeur2020multi} explore multimodal fusion techniques to fuse all the information extracted from a video using multiple pretrained `experts'. While these methods focus on the addition of video-side information, a supervisory signal can also be obtained by looking with more detail at the text. \cite{Chen_2020_CVPR} create a semantic role graph of the caption and aligns to each node a learned representation of the clip-level descriptor. \cite{wray2019fine} extract verbs and nouns from the caption and uses them to learn Part-of-Speech-specific embedding spaces. \cite{patrick2020support} introduce a generative cross-captioning task, using the batched videos as a support set. Recently \cite{croitoru2021teachtext} distil information from multiple pretrained text experts. 
A different trend involves heavy pretraining steps \cite{dzabraev2021mdmmt,lei2021less,bain2021frozen,liu2021hit}, followed by finetuning for downstream tasks. %\rev{To do so, Transformers \cite{vaswani2017attention} are often employed, \eg to model the interactions between the modalities available for the video \cite{gabeur2020multi,dzabraev2021mdmmt}, or to jointly model video and text \cite{lei2021less}.}{} 
Moreover, the addition of image-text datasets as part of the pretraining step, showed significant improvements when dealing with video-related tasks \cite{lei2021less,bain2021frozen}. While these methods achieve impressive results, they rely heavily on the data, are expensive to train, and are not designed for the nature of the problem.

Due to the unavailability of groundtruth relevance values which can inform about the optimal ranking list to a given query, the video retrieval community focused on rank-unaware metrics such as the recall rates or the median rank. Contrastive losses greatly improve these metrics since they reduce the distance between the visual descriptor and the linguistic one and thus increase its similarity, making it possible to retrieve it before the negative descriptors.
But multiple descriptions can be equally or partially relevant for the same video (and vice versa), thus more complex and rich metrics, such as the nDCG, are needed to accurately evaluate a retrieval system \cite{wray2021semantic}. To do so, a way to determine how relevant an item is to a query must be available. To avoid the need for manual and costly annotation, \cite{damen2020rescaling} proposes to use a relevance function defined in terms of the noun and verb classes present in the caption (more details in Sec.~\ref{sec:relevance}).

\textbf{Learning a joint embedding space.} Common approaches for text-video retrieval learn a joint embedding space by means of a contrastive loss \cite{hadsell2006dimensionality,schroff2015facenet} which, during training, puts semantically similar items (\eg a video and a caption describing its contents) closer in the embedding space, while dissimilar items are pushed away. While groundtruth associations (\ie positive pairs, such as a video and its caption) are known from the dataset, the negative examples (such as a different video) have to be sampled, or `mined', given that the amount of possible tuples scales exponentially with the dataset size, \eg cubically with triplets.
Multiple techniques have been proposed including: offline mining, which randomly samples a fixed number of tuples and repeats the process multiple times during training; online mining, which uses the negatives inside the mini-batch by considering all the non-groundtruth pairs, or only hard \cite{hermans2017defense,xuan2020hard} or semi-hard negatives \cite{schroff2015facenet}. Recent research also found relevant signal while mining positive samples, \eg easy \cite{xuan2020improved} or hard positives \cite{hermans2017defense}. In our paper, we focus on triplets as they are a popular margin-based contrastive loss, but it can be extended to other techniques, \eg to quadruplets \cite{chen2017beyond}. Moreover, we experiment with two different and opposite techniques: offline mining with random sampling and online mining with hard negatives, and show the advantages of the relevance-based margin in both cases.

\textbf{Margin in contrastive losses.} \rev{}{Most of the approaches involving contrastive losses are based on maximum-margin losses (e.g. \cite{hadsell2006dimensionality}). Although the margin is usually fixed, variable or adaptive solutions for it have been explored in different fields. For person re-identification, \cite{cheng2016person} suggest using two different (but fixed) margins for inter- and intra-class constraints, whereas \cite{zhang2019learning} propose to monotonically increase the margin during the training process. \cite{hu2018cvm} use a `soft margin' to improve recommender systems, that is they remove the fixed margin and use (a soft version of) the distance between positive and negative pairs as the loss. \cite{li2020symmetric} augment the bidirectional contrastive loss by also summing the margin to the loss objective, to optimize it during the training process. For text-image retrieval, \cite{semedo2019cross} propose a scheduled adaptive margin which starts from a fixed value and gradually changes during the training process both to integrate inter-category similarity-based correlations and to preserve the category clusters formed during the initial phases of the training. Recently, for cross-modal video retrieval \cite{he2021improving} proposed an adaptive margin proportional to the similarity of the representations computed for the negative pair, both in terms of `static' (pretrained, frozen) models, which provide initial supervision, and `dynamic' (trained with the task) models, which provide supervision in later stages of the training. % {\color{red}maybe expand about what static (=fixed margin?) and dynamic (?) refers to}. 
Differently from all these works, we propose a margin which is proportional to the relevance value of the queries involved in the triplet, effectively using the semantic knowledge during training. }

\textbf{Optimization of nDCG.}
Considering that visual-textual datasets usually lack relevance grades, rank-unaware metrics are one of the preferred ways to measure progress in the video retrieval community. Yet given a video, multiple captions can be used to describe its contents. To capture the difference in the ranking list when binary relevance (\ie a caption is either relevant or irrelevant to a video) is considered, mAP is preferred to the recall rates. Furthermore, finer-grained relevance grades could be also available (\ie a caption can be relevant to a video to some degree), in which case the DCG (or its normalized version, the nDCG) is chosen. %To evaluate the retrieval system we use nDCG and mAP (for a formal definition, see Sec.~\ref{sec:relevance}), which are based on the ranking induced by the model.
But, optimizing these metrics during training clashes with gradient-based optimization methods because ranks are not differentiable with respect to the learnable parameters, \eg the nDCG of a list of items to a given query is normalized using the optimal ranking list, which is computed by \textit{sorting} with respect to the relevance values. 

Surrogate losses are used to partially address this problem, which can be categorized into: pointwise (\eg regression loss \cite{cossock2008statistical}), which compare predicted and optimal rank of one item at a time; pairwise (\eg RankNet \cite{burges2005learning}), which deal with pairs of items and relative ordering; listwise approaches (\eg LambdaRank \cite{burges2006learning}), which work on full list of items. Note that the triplet loss \cite{schroff2015facenet} can be seen as a `triplet-wise' surrogate loss. Since these surrogate losses are loosely connected to downstream metrics, there is also an active research field which directly optimizes retrieval metrics by deriving a relaxation of the sorting operator which has well-defined gradients, \eg \cite{grover2018stochastic,cuturi2019differentiable,blondel2020fast}.

Considering its widespread usage for video retrieval, we consider the triplet loss an optimal candidate for our relevance-based margin, and show it can lead to higher quality ranking lists.

%% file: sub/3_method.tex
\section{Relevance-based margin\label{sec:fm}}
In Sec.~\ref{sec:relevance} we define the relevance function $\rel$ and the metrics used during evaluation. In Sec.~\ref{sec:rel_margin} we describe how we change the margin in the contrastive loss to make it dependent on $\rel$. Finally, Sec.~\ref{sec:methods} details the three methods on which we test our technique.
\subsection{Semantic classes and relevance\label{sec:relevance}}
Given a video clip, multiple natural language descriptions may fully capture its visual contents, and vice versa. Hence, if a user looks for videos about `cooking a pizza', an intelligent video retrieval system should retrieve all the videos which show how to cook a pizza, and show them all before (\ie rank them higher than) those that show the baking of a `focaccia'. Similarly, videos about `fried potatoes' should be ranked even lower, given how dissimilar they are when compared to the user query. As a consequence, the automatic evaluation of the quality of a ranking list requires a function which considers `focaccia' more relevant than `potatoes' when compared with `pizza', as well as the cooking technique (`bake' versus `fry'). To avoid the need for costly manual annotation which requires human assessments using a predefined set of grades, \cite{damen2020rescaling} introduces a relevance function $\rel$ defined as:
\begin{equation}\label{eq:relevance}
    \rel(x_i, x_j) = \frac{1}{2} \bigg(\frac{\vert x_i^V \cap x_j^V \vert}{\vert x_i^V \cup x_j^V \vert} + \frac{\vert x_i^N \cap x_j^N \vert}{\vert x_i^N \cup x_j^N \vert} \bigg)
\end{equation}
where $x_k^V$ and $x_k^N$ denote the sets of verb and noun classes found in the $k$-th caption.
%$\cdot^V$ and $\cdot^N$ represent the sets of verb and noun classes found in the $i$-th and $j$-th caption. 
This can be extended to videos by considering the associated description. By defining the relevance as in Eq.~\ref{eq:relevance}, $x_i$ is highly relevant to $x_j$ if they share similar noun and verb classes.
%item,\swathi{maybe write as i-th and j-th videos. then define or explain the relevance function or what it means to have a high value for two captions and vice versa} meaning that when $x_i$ or $x_j$ are videos the associated captions are used in their place. 
We refer to `classes' because we do not want to consider synonyms (\eg `pick up' and `take', or `drop' and `put down') as different items which need to be separated, hence each class will contain tokens with a similar meaning. % belonging to the same class\swathi{we are defining 'classes' with 'class'. maybe a better sentence}. 
In some datasets, this class knowledge may be already available, but several other datasets do not provide it. To automatically compute them, a pipeline made of a PoS-tagger (\eg with spaCy), followed by WordNet \cite{miller1995wordnet} and the Lesk algorithm \cite{lesk1986automatic} can be used, as in \cite{wray2021semantic}. 

To evaluate a video retrieval system, we use two metrics which are commonly used in Information Retrieval, which are the Mean Average Precision (mAP \cite{baeza1999modern}) and the Normalized Discounted Cumulative Gain (nDCG \cite{jarvelin2002cumulated}), as recently proposed in \cite{wray2021semantic}. 
%\swathi{as recently proposed in...}. 
The mAP is defined as the mean of the Average Precision (AP) with respect to all the queries. For a given query $q$, AP can be defined as:
\begin{equation}
    AP(q) = \frac{\sum_{k=1}^N P(k) \cdot r(k)}{N_r}    
\end{equation}
where $N$ is the number of items (both relevant and irrelevant) in the ranking list, $P(k)$ is the Precision at k \cite{baeza1999modern}, $r(k)$ is an indicator function which tells whether the $k$-th item is relevant or not, and $N_r$ is the total number of relevant items. The mAP allows to grasp with a single number the area under the Precision-Recall curve. But this metric requires binary relevance values, thereby requiring the introduction of a threshold below which items are considered irrelevant (and relevant otherwise). %\swathi{split into two sentences: map requires binary relevance. this needs the introduction of a threshold...} 
For mAP, we consider $k$ to be relevant to $q$ only when $\rel(q, x_k)=1$ as is done in \cite{damen2020rescaling} (hence, for mAP $N_r = \vert \{x_i \,\vert\, \rel(q, x_i) = 1 \}\vert$). On the other hand, nDCG makes use of non-binary relevance values, allowing it to grasp finer details (and errors) of the ranking list. Given a query $q$ and a list of items $K = \{x_i \}$, it is defined as
\begin{equation}
    nDCG(q, K) = \frac{DCG(q, K)}{IDCG(q, K)}
\end{equation}
where IDCG is the optimal DCG value obtained when the ranking list follows a descending order of relevance values. We define DCG as in \cite{jarvelin2002cumulated,damen2020rescaling}:
\begin{equation}
    DCG(q, K) = \sum_{k=1}^{N_r} \frac{\rel(q, x_k)}{\text{log}_2 (k+1)}
\end{equation}
%\swathi{how is $K$ used in the equation? $k$ is a number as per eqn4. how can it be used in eqn 1 to compute relevance?}
where $x_k$ is the $k$-th item in the list $K$, and we only consider the first $N_r$ items in the ranking list. Note that $N_r = \vert \{x_i \,\vert\, \rel(q, x_i) > 0 \}\vert$.

\subsection{Contrastive loss with relevance-based margin\label{sec:rel_margin}}
To learn a joint text-video embedding space, various contrastive (or ranking) losses have been proposed (see Sec.~\ref{sec:rw}). In our work we consider a contrastive term based on the triplet loss defined as: 
\begin{equation}
    \mathcal{L} = [m + s(a, n) - s(a, p)]_+
    \label{eq:contrastive}
\end{equation}

\noindent where $[\cdot]_+ = \text{max}(0, \cdot)$, $m$ is interpreted as a separation margin, $s(\cdot, \cdot)$ is a similarity metric (\eg cosine similarity), whereas $a$, $n$, and $p$ represent respectively the embedding of the $a$nchor, $n$egative, and $p$ositive item. Eq.~\ref{eq:contrastive} provides a positive loss when the margin $m$ between the positive pair $(a, p)$ and the negative one $(a, n)$ is violated, %\swathi{this sentence is not clear} 
\ie $s(a, p) - s(a, n) < m$. The loss may be cross-modal, \ie $n$, $p$ from one modality (\eg video) and $a$ from the opposite one (\eg text), or within-modal, \ie $a$, $p$, $n$ are all from the same modality. Furthermore, the optimal $m$ is not known beforehand and should be treated as an hyper-parameter which can affect the performance. Thus, it should be tuned on the validation set. % to obtain optimal performance. %\swathi{maybe mention here that m is usually found through hyperparameter tuning}
%We validate our approach using three video-text retrieval methods, MME \cite{wray2019fine}, JPoSE \cite{wray2019fine}, and HGR \cite{Chen_2020_CVPR}. MME and JPoSE perform offline triplet mining, while HGR performs online triplet mining. More details about the methods can be found in their respective papers. %To mine the triplets, we validate our approach using both offline (for MME and JPoSE \cite{wray2019fine}) and online mining with hard negatives (for HGR \cite{Chen_2020_CVPR}). The formal definition of all the three losses is defined in the respective papers. 
%In the former, the training loss will be defined as
%\begin{equation}
%\begin{split}
%    \mathcal{L_{off}} &= \sum_{(a,p,n) \in \mathcal{T}_{t,v}} [m + s(f(a), g(n)) - s(f(a), g(p))]_+ + \\
%    &+ \sum_{(a,p,n) \in \mathcal{T}_{v,t}} [m + s(f(a), g(n)) - s(f(a), g(p))]_+ + \\
%    &+ \sum_{(a,p,n) \in \mathcal{T}_{v,v}} [m + s(f(a), f(n)) - s(f(a), f(p))]_+ + \\
%    &+ \sum_{(a,p,n) \in \mathcal{T}_{t,t}} [m + s(g(a), g(n)) - s(g(a), g(p))]_+ +
%\end{split}
%\end{equation}
%where $\mathcal{T}_{v,t}$, $\mathcal{T}_{t,v}$, $\mathcal{T}_{v,v}$, and $\mathcal{T}_{t,t}$ are the sets of mined triplets. 

During training, all the items which are not from the positive pair $(a, p)$ are pushed away until they are separated by a margin of $m$, as shown in Fig.~\ref{fig:method}. Although effective and widely used in the literature, Eq.~\ref{eq:contrastive} ignores that multiple items may be completely or partially relevant to the same query, and treats all the items which are not from the groundtruth pair as equally irrelevant. Thus the retrieval system might not be able to distinguish between the many relevance levels which can exist between an item and a query. %Not being able to distinguish between the many relevance levels which can exist between different items and queries leads to retrieval systems which are not able to make such a distinction.\swathi{not clear, what distinction?} Yet, when evaluating a retrieval system, multiple indicators should be assessed, including the `quality' of the ranking list.\swathi{not sure if this sentence is needed here. no solutions are proposed. moreover next sentence says "to address this". it will be confusing what we are trying to address: not considering different relevance levels or ranking list evaluation} 

To address this, we propose a relevance-based margin instead of a fixed margin. In our work, we aim at defining $m$ in terms of the relevance function $\rel$. In particular, we update Eq.~\ref{eq:contrastive} as follows:
\begin{equation}
    \mathcal{L} = [\Delta_{a,p,n} + s(a, n) - s(a, p)]_+
    \label{eq:rel_contrastive}
\end{equation}
\noindent where:
\begin{equation}
    \begin{split}
    \Delta_{a,p,n} &= R(a, p) - R(a, n) \\
    &= 1 - R(a, n)
    \end{split}
\end{equation}
since we consider the groundtruth pair to be maximally relevant, \ie $R(a, p) = 1$. The relevance-based margin keeps $\mathcal{L}$ positive until $s(a, p)$ and $s(a, n)$ are separated by a margin which is proportional to their relevance values, thus separating irrelevant items more than those which have a positive relevance. This is illustrated in Fig.~\ref{fig:method} on the lower right. Note that this term is not bound to the network state and can thus be applied both to offline and online mining techniques.

\subsection{Methods \label{sec:methods}}
Given a dataset $D = \{ (v_i, q_i) \}$ of video-caption pairs, %\swathi{not sure if we have to consider videos with multiple captions} 
we strive to learn optimal weights for two embedding functions $f: \R^{f_v} \rightarrow \R^d$ and $g: \R^{f_q} \rightarrow \R^d$ such that $f(v_i)$ and $g(q_i)$ are close in the $d$-dimensional joint embedding space. Here $f_v$ and $f_q$ represent the dimension of the video and caption descriptors. %\swathi{what are $d$, $f_v$ and $f_q$?} 
To parameterize $f$ and $g$ we consider the following methods: \textbf{MME} is a baseline from \cite{wray2019fine} which learns one embedding function for each of the two modalities, video and text. %\textbf{MoEE \cite{miech2018learning}} embeds multiple representations of the video data by using pretrained `experts', which are then combined using the caption to weight each contribution.
In \noindent\textbf{JPoSE \cite{wray2019fine}}, the captions are processed in order to obtain a single sentence-level descriptor and multiple descriptors restricted to specific Part-of-Speech (PoS) tags, \eg nouns and verbs. Then, two functions are learned for each of these embedding spaces using both cross-modal and intra-modal contrastive terms for the sentence-level, as well as for the PoS-level. \textbf{HGR \cite{Chen_2020_CVPR}} structures the learning at multiple levels (global event, local actions, and local entities) which are obtained by computing a semantic role graph for each of the captions. Then a graph convolutional network is used to learn compositional semantics of the caption based on the local components, \ie full sentence, verbs, and noun phrases. %computes a semantic role graph for each of the captions, and structures the learning at multiple levels by exploiting a graph convolutional network to learn compositional semantics of the caption based on the local components.\swathi{not clear what multiple levels and local components} 

We choose these three methods because they provide incrementally structured approaches to deal with video and language data, starting from a simpler MLP-based network to a graph-based approach. \rev{}{Moreover, JPoSE represents the state-of-the-art for EPIC-Kitchens-100 (measured through nDCG and mAP), which is the main dataset under consideration. Finally, by selecting them we can validate our approach on both offline (MME and JPoSE) and online (HGR) mining techniques.} We thus proceed to show the generality and effectiveness of the proposed relevance-based margin by empirically validating on two different datasets.%\textcolor{red}{swathi: maybe also mention they use different negative mining: offline and online} % \sergio{motivate explicitely why the selection of these 3 methods: to show generality of our relevance different method behaviors and accurate results in datasets or blabla...}

%% file: sub/4_results.tex
\section{Experiments\label{sec:exp}}
After the introduction of the experimental setting in Sec.~\ref{sec:expset}, we show in Sec.~\ref{exp:ek100_base} how the proposed relevance-based margin helps to achieve better nDCG and mAP on EPIC-Kitchens-100 and YouCook2. Then, in Sec.~\ref{exp:abl} we perform several ablation studies. First we show that even by carefully tuning the fixed margin, the proposed technique consistently achieves better results without the need to tune it. Secondly, we also evaluate its robustness by ablating the loss function and the modalities used in JPoSE. Finally in Sec.~\ref{sec:exp_qual} we \rev{}{analyze the distribution of the margin values during training and some} video-to-text examples from the testing set.
%\sergio{if we have space it would be nice to use few lines here to explain what this section contains: first we will evaluate for X and then Y ... otherwise there is a lot of uncertainly for the reader about what will happen next :) }

\subsection{Experimental setting\label{sec:expset}}
\textbf{Datasets.} We focus our experimental setting on two challenging video and language datasets: \rev{}{the recently released} EPIC-Kitchens-100 \cite{damen2020rescaling} and YouCook2 \cite{zhou2018towards}. For the retrieval challenge, the former comprises 67217 egocentric clips for training and 9668 for evaluation. \rev{}{It is also the largest dataset for video retrieval in the egocentric setting.} %\textcolor{red}{swathi:mention that it is most recent and has largest number of videos} 
Moreover, it also provides semantic annotations for each of the captions, by covering 300 noun and 97 verb classes. The latter provides a lower amount of training clips (10337) but still offers a challenging evaluation set with 3492 clips. While semantic annotations are not available for YouCook2 they can be computed using WordNet and the Lesk algorithm, as described in Sec.~\ref{sec:relevance}. Furthermore, as both EPIC-Kitchens-100 and YouCook2 share the kitchens domain, the class knowledge of the former can also be used for the latter \cite{wray2021semantic}.

\textbf{Implementation details.}
For EPIC-Kitchens-100 we use the TBN \cite{kazakos2019epic} features from the dataset provider comprising of 25 uniformly sampled RGB, flow, and audio feature vectors for each clip. For YouCook2 we use ImageNet-pretrained ResNet-152 features from the VALUE benchmark \cite{li2021value}. For the three methods we use the open source codebases provided in the respective papers and follow their hyper-parameter setting. We release our code and models on GitHub to support reproducibility.

\subsection{Relevance-based margin results\label{exp:ek100_base}}
\textbf{EPIC-Kitchens-100.} To validate the effectiveness of the proposed relevance-based margin, we explore three methods (MME, JPoSE, and HGR as described in Sec.~\ref{sec:methods}) on EPIC-Kitchens-100. In Tab.~\ref{tab:ndcg_methods} we report nDCG and mAP values, averaged between text-to-video and video-to-text. In all three cases, we observe a large improvement in both metrics, showing that the relevance-based margin works on very different models. It also works well with both offline mining with randomly sampled triplets (for MME and JPoSE), and online mining with hard negatives (for HGR): by using the relevance-based margin, MME gains +1.1 nDCG and +0.7 mAP, JPoSE +2.7 nDCG and +1.8 mAP, and finally HGR obtains +18 nDCG and +9.6 mAP. \rev{}{Such a large improvement is possibly due to how the triplets are sampled: in JPoSE, the negatives do not share the verb class of the anchor, leading to a relevance lower than 0.5; but, there is not such a guarantee in HGR, since batches are formed randomly. Hence, by employing a relevance-based margin in HGR we automatically deal with situations in which the negatives have a considerable relevance and adapt the margin accordingly. Finally, in \new{App.~\ref{sup:leaderboard}} we report the public leaderboard for the retrieval challenge, confirming the improvement we observe over current state-of-the-art methods.}

\begin{table}[]
    \centering
    \begin{tabular}{c|c|c@{\hskip -1mm}cc@{\hskip -0.01mm}c} \hline
        Method & rel-$\Delta$ & nDCG & & mAP & \\ \hline
        MME & & 48.5 & & 38.5 & \\
        MME & \checkmark & 49.6 & {\bf\color{darkgreen}\footnotesize$\uparrow$1.1} & 39.2 & {\bf\color{darkgreen}\footnotesize$\uparrow$0.7} \\ \hline
        JPoSE & & \underline{53.5} & & 44.0 & \\
        JPoSE & \checkmark & \textbf{56.2} & {\bf\color{darkgreen}\footnotesize$\uparrow$2.7} & \textbf{45.8} & {\bf\color{darkgreen}\footnotesize$\uparrow$1.8} \\ \hline
        HGR & & 32.2 & & 36.0 &\\ 
        HGR & \checkmark & 50.2 & {\bf\color{darkgreen}\footnotesize$\uparrow$18} & \underline{45.6} & {\bf\color{darkgreen}\footnotesize$\uparrow$9.6} \\ \hline
    \end{tabular}
    \caption{nDCG and mAP results on EPIC-Kitchens-100 with three different methods, using TBN (RGB, Flow, Audio) features. \rev{}{We report in bold the best results (and underline the second best). With ``$\uparrow$X'' we represent an improvement of X when compared to the above result.}}
    \label{tab:ndcg_methods}
\end{table}

\textbf{YouCook2.} In the previous experiment we used the class knowledge which accompanies the dataset. But, by computing synsets knowledge in a similar way to what is done in EPIC-Kitchens-100, the proposed relevance-based margin can still successfully help the training process. This setting poses two additional challenges: first of all, in EPIC-Kitchens-100 most of the captions follow a precise structure, \ie they contain a verb and a noun, which is not the case when dealing with other datasets, where free-form descriptions are often adopted. This may make it more difficult for the PoS-tagger to correctly tag the words. Secondly, there may be words which are put in the wrong category by WordNet. 

For this dataset, we use the same class knowledge used in EPIC-Kitchens-100, as it transfers well across both datasets since they share the cooking domain \cite{wray2021semantic}, and for words which do not appear in any class, a new singleton class is created. 

%We evaluate the proposed approach on YouCook2 using HGR with ImageNet-pretrained ResNet-152 features from the VALUE benchmark \cite{li2021value}. 
In Tab.~\ref{tab:res_snd_dataset} we report the nDCG and mAP values obtained with MME, JPoSE, and HGR. From the table, one can see that even in this different setting the relevance-based margin is able to provide useful information to the model. For example, the addition of the proposed technique in HGR leads to a gain of +5.5 nDCG and +3.1 mAP when compared to the results obtained with a fixed margin. %\textcolor{red}{swathi: any reason why hgr improvement is large?}

\begin{table}[]
    \centering
    \begin{tabular}{c|c|c@{\hskip -0.85mm}cc@{\hskip -0.01mm}c} \hline
        Method & rel-$\Delta$ & nDCG & & mAP & \\ \hline
        \multirow{2}{*}{MME} & & 46.9 & & 19.3 \\
        %MME & \checkmark & 46.6{\bf\color{orange}\footnotesize$\downarrow0.3$} & & 19.2{\bf\color{orange}\footnotesize$\downarrow0.1$} \\
        & \checkmark & 47.3 & {\bf\color{darkgreen}\footnotesize$\uparrow$0.4} & 19.5 & {\bf\color{darkgreen}\footnotesize$\uparrow$0.2} \\ \hline
        \multirow{2}{*}{JPoSE} & & \underline{49.6} & & 20.5 \\
        %JPoSE & \checkmark & 50.7{\bf\color{darkdarkgreen}\footnotesize$\uparrow1.1$} & & 22.1{\bf\color{darkdarkgreen}\footnotesize$\uparrow1.6$} \\
        & \checkmark & \textbf{50.4} & {\bf\color{darkgreen}\footnotesize$\uparrow$0.8} & 21.5 & {\bf\color{darkgreen}\footnotesize$\uparrow$1.0} \\ \hline
        \multirow{2}{*}{HGR} & & 41.0 & & \underline{23.0} & \\
        & \checkmark & 46.5 & {\bf\color{darkgreen}\footnotesize$\uparrow$5.5} & \textbf{26.1} & {\bf\color{darkgreen}\footnotesize$\uparrow$3.1} \\ \hline
        %VATEX &  &  &  \\
        %VATEX & \checkmark &  &  \\ \hline
    \end{tabular}
    \caption{nDCG and mAP using MME, JPoSE, and HGR on YouCook2. We use ResNet-152 (pretrained on ImageNet) features from the VALUE benchmark \cite{li2021value}.} %; for the latter, we use the S3D features provided by the authors.}
    \label{tab:res_snd_dataset}
\end{table}

\subsection{Ablation studies\label{exp:abl}}
We perform the ablation studies on EPIC-Kitchens-100 using JPoSE.

\begin{figure}
    \centering
    \includegraphics[width=\linewidth]{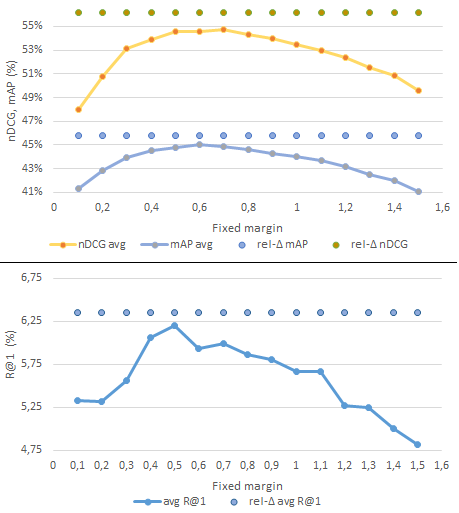}
    \caption{Using JPoSE on EPIC-Kitchens-100, we show how changing the fixed margin in the loss function affects the performance, measured through nDCG and mAP in the upper figure, and \rev{}{average} R@1 in the lower one. For reference, we also plot disconnected dots to show the performance when we use the proposed relevance-based margin. Notice that the optimal fixed-margin hyper-parameter would not be known in a testing scenario; it would need to be estimated through an expensive hyper-parameter search on a validation set.} % we get 56.2\% nDCG, 45.8\% mAP, 5.9\% text-to-video R@1, and 6.7\% video-to-text R@1.\swathi{can we add the relevance-based margin score in the graph (perhaps as a line spanning the whole x axis). this will make it easier to see the improvements. the graph can be split into two for 1) map and ndcg; and 2) R@1}}
    \label{fig:vary_fixed}
\end{figure}

\textbf{Varying the fixed margin.} In Sec.~\ref{exp:ek100_base} we show that the proposed relevance-based margin leads to improved nDCG and mAP on both EPIC-Kitchens-100 and YouCook2. But, what if one would only need to carefully tune the fixed margin to obtain similar results? To answer to this question, we focus on JPoSE and vary the fixed margin $\Delta$ in $\{0.1, 0.2, \dots, 1.5\}$ (default value used in JPoSE is 1.0). We keep the rest of the hyper-parameter setting as in \cite{wray2019fine,damen2020rescaling} and use the officially provided TBN features. We plot in Fig.~\ref{fig:vary_fixed} nDCG, mAP, \rev{}{average} R@1 for each of the tested margins. While small margins lead to worse results overall, it can be seen that increasing the margin does not improve significantly neither the nDCG nor the mAP. Moreover, the recall rates are affected only marginally as well. When compared to the performance shown by the adoption of the relevance-based margin, it can be observed that our technique manages to achieve higher nDCG and mAP values, while also keeping similar recall rates (\rev{}{on average, 6.3\%} R@1). Finally, it is worth noticing that by using the relevance-based margin we are released from the margin hyper-parameter: this is also of practical importance, because by using a fixed margin its optimal value is not known in a testing scenario, hence one would also need to perform an expensive search on the validation set in order to achieve better performance.

\textbf{Losses ablation.} A peculiarity of JPoSE is that it uses multiple contrastive loss terms to learn both global- and PoS-restricted joint embedding spaces. To do so, the authors employ a global loss and a PoS-level loss, both in the cross- and within-modality settings. We perform an ablation study in Tab.~\ref{tab:jpose_abl} to give evidence that the relevance-based margin can be helpful even when restricting the amount of loss terms used. Note that when applying the technique to the PoS-level terms (\eg verbs) we consider the term for the opposite PoS (\eg nouns) in Eq.~\ref{eq:relevance} to be 1. As shown in Tab.~\ref{tab:jpose_abl}, the adoption of the relevance-based margin leads to an improvement of +1.6 nDCG and +1.2 mAP when using only the cross-modal global-level loss terms, whereas +2.8 nDCG and +1.9 mAP are gained when also adding the cross-modal PoS-level terms. 

\begin{table}[]
    \centering
    \begin{tabular}{c|c@{\hskip 1mm}c@{\hskip 1.2mm}c|c@{\hskip -0.85mm}c@{\hskip 1.5mm}c@{\hskip -0.01mm}c} \hline
        & \multicolumn{2}{c}{cross-} & within- & & \\
        rel-$\Delta$ & glob & PoS & glob+PoS & nDCG & & mAP &\\ \hline
         & \checkmark & & & 53.1 & & 43.3 &\\
        \checkmark & \checkmark & & & \underline{54.7} &{\bf\color{darkgreen}\footnotesize$\uparrow$1.6} & 44.5& {\bf\color{darkgreen}\footnotesize$\uparrow$1.2} \\ \hline
         & \checkmark & \checkmark & & 53.4& & 43.7& \\
        \checkmark & \checkmark & \checkmark & & \textbf{56.2} &{\bf\color{darkgreen}\footnotesize$\uparrow$2.8} & \underline{45.6}& {\bf\color{darkgreen}\footnotesize$\uparrow$1.9} \\ \hline
         & \checkmark & \checkmark & \checkmark & 53.5 & &44.0 &\\
        \checkmark & \checkmark & \checkmark & \checkmark & \textbf{56.2} &{\bf\color{darkgreen}\footnotesize$\uparrow$2.7} & \textbf{45.8}& {\bf\color{darkgreen}\footnotesize$\uparrow$1.8} \\\hline
    \end{tabular}
    \caption{nDCG and mAP using JPoSE on EPIC-Kitchens-100. During training, JPoSE considers both cross- and within-modality contrastive losses, both at sentence- and PoS-level. Applying the relevance-based margin helps at each level.}
    \label{tab:jpose_abl}
\end{table}

\textbf{Modalities ablation.} For EPIC-Kitchens-100 we have RGB, flow, and audio features. To show that the improvements obtained when applying the relevance-based margin are not due to the model accessing multiple modalities related to the video, we perform another ablation in Tab.~\ref{tab:rgb_flow_audio_abl} by considering RGB-only and RGB+flow features. In both cases the proposed technique shows its usefulness. In particular, by employing the relevance-based margin we observe +1.6 nDCG and +1.6 mAP when using RGB-only, +2.9 nDCG and +1.8 mAP when using both RGB and flow, and +2.7 nDCG and +1.8 mAP when adopting all the three modalities. %Moreover, it can be seen that the relevance-based margin is particularly effective when using RGB-only (+1.6 nDCG and +1.6 mAP) or both RGB and flow features (+2.9 nDCG and +1.8 mAP). On the other hand, audio features provide helpful information to the model, but the interaction with the relevance-based margin is not as fruitful as with the previous modalities.\swathi{similar gain is obtained! (2.7 and 1.8)} \sergio{yes I also see that large improvement when tri-modal.}

\begin{table}[]
    \centering
    \begin{tabular}{l|c|c@{\hskip -0.85mm}cc@{\hskip -0.005mm}c} \hline
        Modalities & rel-$\Delta$ & nDCG & & mAP &  \\ \hline
        \multirow{2}{*}{RGB} & & 36.8 & & 28.8 & \\
         & \checkmark & 38.4 & {\bf\color{darkgreen}\footnotesize$\uparrow$1.6} & 30.4 & {\bf\color{darkgreen}\footnotesize$\uparrow$1.6} \\ \hline
        \multirow{2}{*}{RGB+Flow} & & 49.6 & & 41.0 & \\
         & \checkmark & 52.5 & {\bf\color{darkgreen}\footnotesize$\uparrow$2.9} & 42.8 & {\bf\color{darkgreen}\footnotesize$\uparrow$1.8} \\ \hline
        \multirow{2}{*}{RGB+Flow+Audio} & & \underline{53.5} & & \underline{44.0} &  \\
         & \checkmark & \textbf{56.2} & {\bf\color{darkgreen}\footnotesize$\uparrow$2.7} & \textbf{45.8} & {\bf\color{darkgreen}\footnotesize$\uparrow$1.8} \\ \hline
    \end{tabular}
    \caption{TBN offers RGB, flow, and audio features. The proposed relevance-based margin interacts with each modality in an incremental way. We use JPoSE on EPIC-Kitchens-100.}
    \label{tab:rgb_flow_audio_abl}
\end{table}

\subsection{Qualitative analysis\label{sec:exp_qual}}
\begin{figure}
    \centering
    \includegraphics[width=\linewidth]{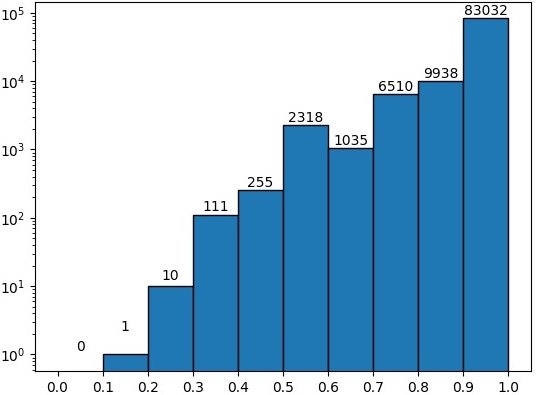}
    \caption{Log-scale distribution of the margins used during training. Over each bin we report the frequency. Numbers refer to one epoch, with 10 triplets sampled for each example (e.g.~with 10337 examples for YouCook2, we end up with around 103k triplets). Although a great part of the triplets are separated with the highest margin (i.e. lowest relevance), around 20k triplets are distanced by various margin values.}%{\color{red}maybe tell that it is log-scale}}
    \label{fig:distrib}
\end{figure}

\rev{}{First of all, the proposed technique leads to variable margins, therefore the distribution of the values may help explaining why we observe such a positive influence on the final performance. In Fig.~\ref{fig:distrib} we plot the frequencies of the margins (with bins of size 0.1) observed during the training of JPoSE on YouCook2, where for each of the training examples 10 triplets are sampled. It can be seen that a great part of the margins used are in the final bin (between 0.9 and 1.0), for which the relevance is quite low since the margin is computed as $\Delta_{a,p,n} = 1 - \rel(a, n)$ (see Eq.~7). In these cases, the margin will be similar to the default case of JPoSE, i.e.~1.0. Yet, around 20\% of the training triplets end up having smaller margins. In these situations, the varying margins help the model achieve better performance by providing a semantic supervision on the structure of the embedding space, since the relevant items are kept at a distance which is proportional to the relevance.} % easier {\color{red}what do you mean by easier?} constraints which can help in pulling the relevant items close to the query.}

\begin{figure*}
    \centering
    \includegraphics[width=.99\linewidth]{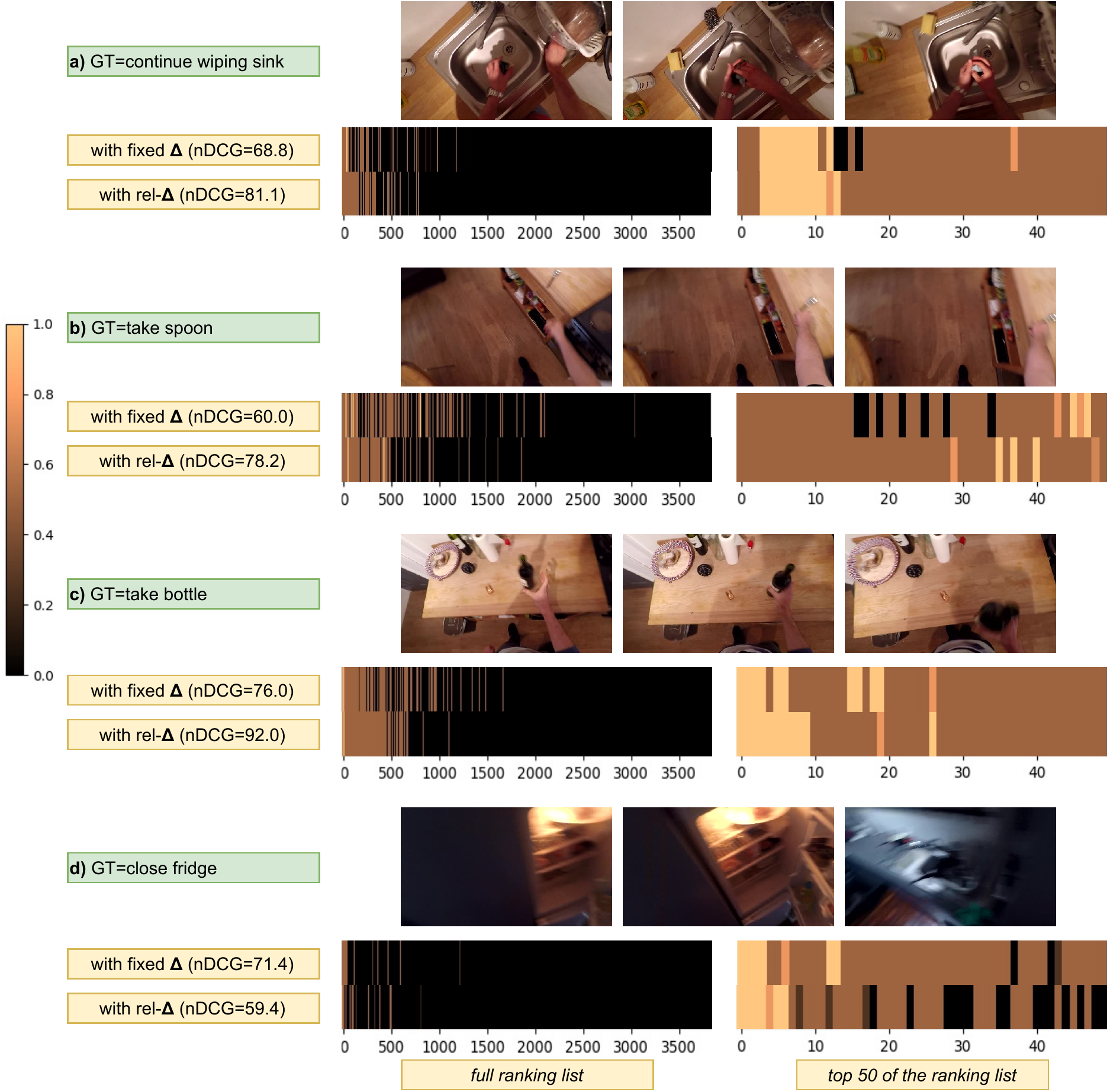}
    \caption{Video-to-text qualitative results on EPIC-Kitchens-100 testing set using JPoSE. For each of the examples we show a few frames and the groundtruth (GT) caption, and we plot both the full ranking list and the top 50 retrieved captions when adopting the fixed margin and then the relevance-based margin. On the left we also visualize the color bar which is used for the relevance (light colors mean high relevance, dark colors low relevance). In particular, Figures a, b, and c are success cases, whereas Figure d represents a failure case.}
    \label{fig:qual}
\end{figure*}

Secondly, in Fig.~\ref{fig:qual} we visualize a few video-to-text examples from the testing set, by plotting for each of them the relevance values of each caption in both the full ranking list and the top 50 retrieved captions. By plotting the full ranking list, it is possible to see that the relevance-based margin helps improving the nDCG, as relevant captions are retrieved first. This can also be seen in the top 50 of Fig.\ref{fig:qual}.a, \ref{fig:qual}.b, and \ref{fig:qual}.c where with the relevance-based margin no irrelevant captions are retrieved and, especially in Fig.~\ref{fig:qual}.c, the ranking is almost ideal. Yet we can still find examples where the proposed technique fails to achieve the expected improvements. In Fig.~\ref{fig:qual}.d, using the relevance-based margin a few irrelevant captions are retrieved, such as `take container' and `take milk container'. This behavior is likely related to the fact that during training captions like `close container' and `close milk container' are relevant (0.5) for a video depicting the action `close fridge', since they share the same verb class. This leads to an increase in the similarity of the respective descriptors. Hence, during inference, also captions like `take container' and `take milk container' might have a significant similarity with the video descriptor of `close fridge'. \rev{}{Further qualitative analysis is available in \new{Appendix \ref{sup:sim_qual}}.}

%\begin{figure*}
%    \centering
%    \includegraphics[width=\linewidth]{images/QualityEval_CVPR22_RelMargin_2.okko.pdf}
%    \caption{Video-to-text qualitative results on EPIC-Kitchens-100 testing set using JPoSE. For each of the examples we show a few frames and the groundtruth (GT) caption, and we plot both the full ranking list and the top 50 retrieved captions when adopting the fixed margin and then the relevance-based margin. We also plot the ideal ranking list. On the left we also visualize the color bar which is used for the relevance (light colors mean high relevance, dark colors low relevance).}
%    \label{fig:qual}
%\end{figure*}

%% file: sub/5_discussion.tex
\section{Conclusions\label{sec:cs}}
Learning a joint embedding space using a margin-based contrastive loss is the dominant approach to deal with text-video retrieval. In the literature it is shown that by using such a framework, competitive performance on rank-unaware metrics, such as the recall rates, can be obtained. Yet, rank-aware metrics, such as the nDCG, need to be taken into account, as multiple descriptions can have numerous levels of relevance to a given query \cite{wray2021semantic}. In this work, we proposed to move away from the fixed margin which is typically used in such a framework, and introduced a relevance-based margin. In particular, we adopted the proposed technique into three increasingly more complex models on two datasets and %\sergio{rewrite in a more simple and readable way: supported by the usage of three varied and increasingly more complex models on two datasets, 
gave empirical evidence that %by employing our proposed technique 
we can easily improve the performance measured through nDCG and mAP.  %(Tab.~\ref{tab:ndcg_methods} and \ref{tab:res_snd_dataset})\oswald{no pointers to figures in concs}. 
Moreover, we showed that even by performing an expensive search of the fixed margin hyper-parameter, it does not reach the same performance as when using the relevance-based margin. 
\rev{}{Furthermore, the proposed technique can also have a positive impact on video retrieval applications as not needing to tune the margin can lead to less GPU hours required to fully train the model.}
Finally, we focused our work on text-video retrieval, but the relevance-based margin can be easily extended to other domains where similar margin-based ranking losses are used, \eg in image retrieval \cite{zhang2020context}. \rev{}{Moreover, we showed the effectiveness of the proposed approach by applying it to loss functions where the margin is explicitly defined and used to separate positive and negative pairs, \eg \cite{schroff2015facenet,chen2017beyond}. Yet, there are also popular loss functions which do not make use of it, such as NCE \cite{gutmann2010noise} and MIL-NCE \cite{miech2020end}. Future work is required to adapt the relevance-based margin to non-margin based loss functions.}

%% file: sub/A_supplementary.tex
\appendix
%In the main paper we showed that adopting a relevance-based margin, in place of the widely used fixed margin, can be beneficial to the performance of the model. Here we further explore some details that were left out due to space constraints. In particular, in Section \ref{sup:leaderboard} we present a comparison to the state-of-the-art methods on EPIC-Kitchens-100 by participating into the public challenge. Secondly, in Section \ref{sup:sim_qual} we present further qualitative analysis where we show how the proposed technique can effectively address some issues which can be observed when employing a fixed margin.

\section{Comparison with the EPIC-Kitchens-100 Challenge leaderboard\label{sup:leaderboard}}
\begin{figure*}
    \centering
    \includegraphics[width=\linewidth]{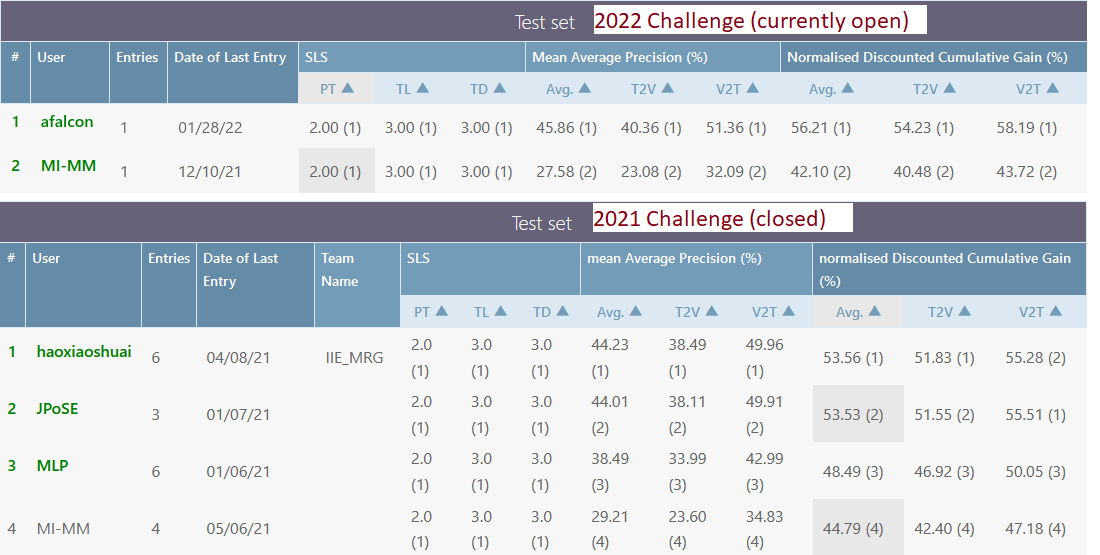}
    \caption{We report the public leaderboard for the EPIC-Kitchens-100 Challenge at time of submission (\textbf{below}), and also the leaderboard for the previous challenge which ended in November 2021 (\textbf{above}). It can be seen that we achieve around +1.6\% mAP and +2.6\% nDCG over the previous best results, achieved by Hao et al. (details in the technical report \cite{Damen2021CHALLENGES}).}
    \label{fig:ek100_leaderboard}
\end{figure*}

The release of the EPIC-Kitchens-100 dataset \cite{damen2020rescaling} was accompanied by a public challenge for the multi-instance retrieval problem (alongside other challenges, e.g. for Action Recognition). To further prove the results we show in Section \ref{sec:exp}, we took part into the challenge by employing the proposed relevance-based margin on the JPoSE method \cite{wray2019fine} (see Section 3). We show the results of both the participants at the time of submission and those that took part into the previous challenge (which ended in November 2021) in Figure \ref{fig:ek100_leaderboard}. The previous best result was obtained by Hao et al. (more details in the technical report \cite{Damen2021CHALLENGES}), which achieved on average 44.23\% mAP and 53.56\% nDCG. As can be seen, we achieve 45.86\% mAP (+1.63\%) and 56.21\% nDCG (+2.65\%).

\section{Qualitative analysis\label{sup:sim_qual}}
\begin{figure*}
    \centering
    \includegraphics[width=\textwidth]{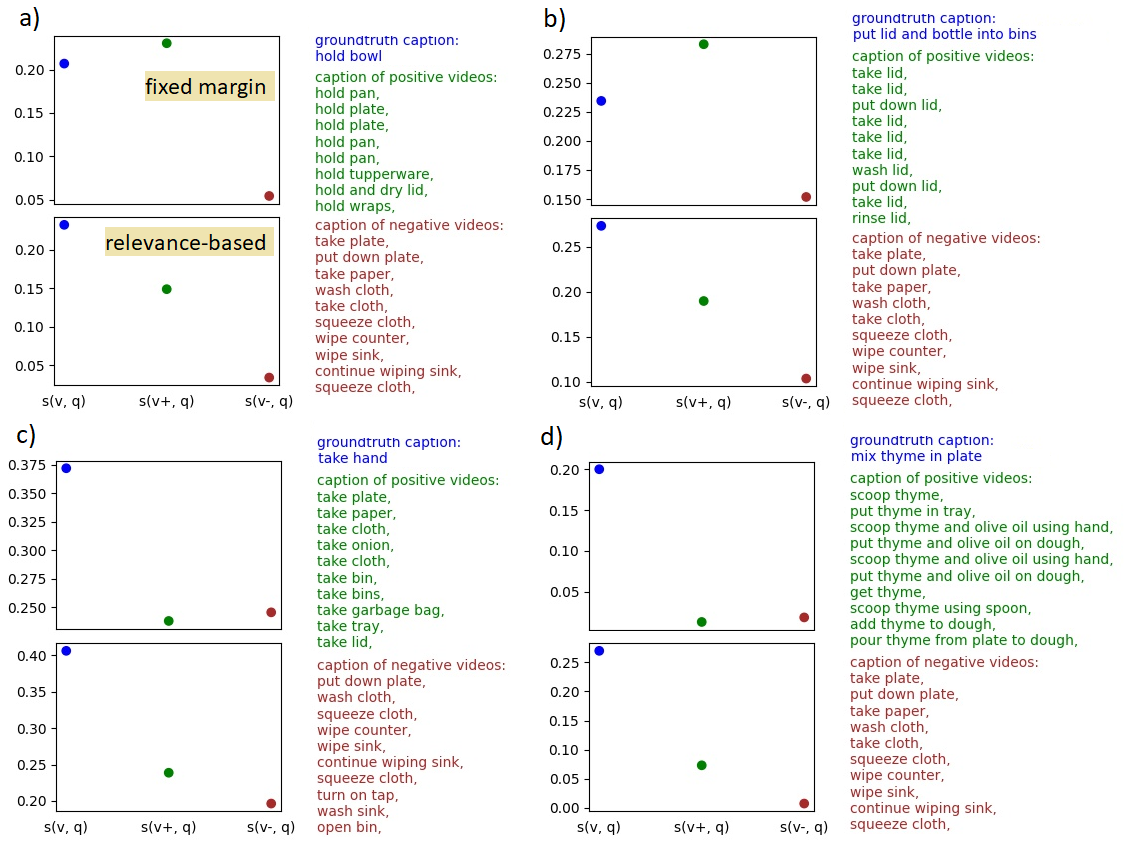}
    \caption{Using JPoSE, we compute a similarity score $s(v, q)$ for the groundtruth pair (colored in blue), $s(v+, q)$ for videos with similar captions (colored in green), and $s(v-, q)$ for videos with dissimilar captions (colored in brown). Note that, when selecting $V_+$, for the examples on the left we change the noun class, whereas on the right we change the verb class. See Sec.~\ref{sup:sim_qual} for more details. The captions of the videos used are reported on the right. Each of the four examples are taken from EPIC-Kitchens-100 testing set and for each of them we report first what happens with fixed margin, then with the proposed relevance-based margin.}
    \label{fig:sim_qual}
\end{figure*}

We further analyze the effectiveness of the proposed technique from a qualitative point of view. To do so, we select three types of information. First of all, we pick a caption and compute its embedding ($q$), pick the corresponding video descriptor ($v$), and compute their similarity $s(v, q)$ through dot product. Then, we look for 10 similar captions (i.e.~different captions which either share the noun or the verb class), pick the corresponding video descriptors indexed by $V_+$, and compute an aggregated similarity value $s(v+, q) = \frac{1}{10} \sum_{v_i \in V_+} s(v_i, q)$. Finally, we also randomly select 10 dissimilar captions (i.e.~sharing neither the verb nor the first noun class), pick their video descriptors, and compute $s(v-, q)$. We compare the results using JPoSE on the testing set of EPIC-Kitchens-100, and report several examples in Figure \ref{fig:sim_qual}. In Figures \ref{fig:sim_qual}.a and \ref{fig:sim_qual}.b the usage of a fixed margin leads to a far too high similarity of the videos in $V_+$ with the query $q$ when compared to its groundtruth video descriptor $v$, which hurts both nDCG, mAP, and the recall rates. In Figures \ref{fig:sim_qual}.c and \ref{fig:sim_qual}.d the videos in $V_-$ and those in $V_+$ are not properly separated, hence wrongly giving the model the impression that they are similarly relevant to the query $q$. In all these cases, adopting a relevance-based margin is a successful strategy to correct these wrong predictions, leading to a more robust model. 

